\documentclass[letterpaper, 10 pt, conference]{ieeeconf}  

\IEEEoverridecommandlockouts                              
\usepackage{mwe}
\usepackage{graphics} 
\usepackage{graphicx}
\usepackage{bm}
\usepackage{siunitx}
\usepackage{relsize}
\usepackage{mathptmx}
\usepackage{amsmath} 
\usepackage{amssymb}  
\usepackage{booktabs}
\usepackage{siunitx}
\usepackage{multirow}
\usepackage{makecell}
\usepackage{hyperref}
\usepackage[usenames, dvipsnames,table,xcdraw]{xcolor}
\usepackage{pifont}

\usepackage[capitalise]{cleveref}
\usepackage{bm}
\usepackage{mathtools}
\usepackage[caption=false,font=footnotesize]{subfig}

\renewcommand{\cdot}{}

\usepackage{changes} 
\usepackage{import}
\usepackage{pgf}          

\definechangesauthor[name={Rohit}, color=blue]{A} 
\definechangesauthor[name={Franek}, color=red]{B} 
\definechangesauthor[name={Felix}, color=green]{C} 
\title{\LARGE \bf Design and Analysis of a Nonlinear Parallel Mechanism Coupling \\
for Dynamic Humanoid Legs}
\author{Rohit Kumar$^{1*}$, Heiner Peters$^{1*}$, Felix Wiebe$^{1}$, Franek Stark$^{1}$, Janne Janzen$^{1}$,  \\
Dennis Mronga$^{1}$, Shubham Vyas$^{1}$, and Frank Kirchner$^{1,2}$ 
\thanks{This work was done in the CoEx (Grant Number 01IW24008), and ActGPT (Grant Number 16IW25002) projects funded by the German Aerospace Center (DLR) with federal funds from the Federal Ministry for Research, Technology and Aeronautics (BMFTR) under the Robotics Institute Germany (RIG)}
\thanks{$^{1}$Robotics Innovation Center, DFKI GmbH, 28359 Bremen, Germany.}%
\thanks{$^{2}$AG Robotik, University of Bremen, 28359 Bremen, Germany.}%
\thanks{$^{*}$Both authors contributed equally to the work}%
\thanks{{Corresponding author}:
    \href{mailto:r.kumar@dfki.de}{r.kumar@dfki.de}} 
}
\begin{document}

\maketitle
\thispagestyle{empty}
\pagestyle{empty}

\begin{abstract}
Humanoids increasingly demand lightweight leg mechanisms for agile, dynamic,
and energy-efficient locomotion. Reducing distal inertia through
proximal actuation has thus become a common design goal. Yet, compact transmissions that
preserve torque capability and efficient force transmission remain hard to
achieve. This paper presents a novel series-parallel leg design combining a knee parallel
mechanism, an ankle parallel mechanism, and a nonlinear knee-ankle coupling where
all actuators placed closer to the hip. As the knee flexes, the
coupling passively assists ankle dorsiflexion for foot clearance and, as it
extends it assists plantarflexion for push-off, redistributing ankle load
through the knee transmission.
We derive the coupling's nonlinear transmission models and validate them in simulation
and on hardware. The design shows that the shared transmission lets both joints
draw on each other's actuator capacity by easing
the knee's peak torque demand under load and the ankle's peak speed
demand during fast motion. 
With the leg's proximal actuator placement, the 
leg achieves an effective reflected mass of $0.37$ kg at the foot
for a $6$ kg leg, demonstrating very low distal leg inertia
for dynamic humanoid robots.

\end{abstract}

\section{Introduction}
\label{sec:intro}

The recent years have seen large scale adoption of humanoid robots. They are becoming increasingly popular 
as they step out from research labs to commercial applications. 
The recent dynamic and agile humanoids are largely
series-parallel hybrid robots where for some joints the actuator is placed
proximally to the joint it drives.
The serial chain helps exploit a larger workspace, while the parallel kinematic 
mechanisms (PKMs) provides high stiffness and high payload capacity for the joints it
actuates. Some prominent examples include commercial platforms such as
Unitree G1, Tesla Optimus, UBTech
Walker, as well as research platforms such as PAL
Kangaroo \cite{MINGOHOFFMAN2024104827}, the Disney bipedal robot
\cite{disney_series_parallel_leg}, the MIT Humanoid \cite{mit_humanoid},
Berkeley Humanoid \cite{berkely_humanoid}, the Tello Leg \cite{tello_leg},
WALKMAN \cite{walkman-iit}, and RH5 \cite{julian_rh5}. 
An extensive survey of such series-parallel hybrid robots is
available in \cite{kumar2020survey}.
Among these platforms, the specific PKM architectures used for the knee and ankle vary considerably.

A common design objective is to reduce the lower leg's inertia. 
Cassie's leg realizes this with the knee and ankle
as a four-bar linkage, with the actuators mounted near the hip but the
ankle remaining local to the lower leg \cite{cassie}. 
LOLA \cite{lola_tum} actuates its 2-DOF ankle through two spatial
slider-crank mechanisms rather than placing motors directly at the ankle
joint, while keeping the knee serial. RH5 \cite{julian_rh5} follows a similar
approach, realizing the ankle and knee as PKMs driven by
linear actuators. The Disney bipedal robot \cite{disney_series_parallel_leg}
combines a 5-bar linkage for the hip and knee with a separate servo mounted
directly at the foot for ankle pitch. 
\cite{lutz_bipetto} illustrates four-bar knee and ankle linkages on the Bipetto robot. 
In each of these designs, the ankle
actuator remains below the knee joint, rather than being relocated closer to
the hip.
\begin{figure}[!t]
	\centering
	\includegraphics[width=0.89\linewidth]{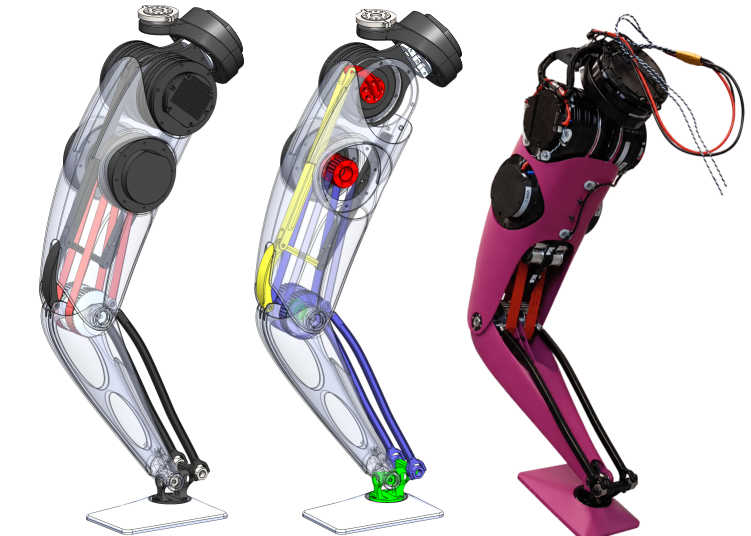}
	\caption{CAD view (left), parallel mechanisms (middle), and real system (right). Knee mechanism is 
	shown in yellow and ankle mechanism is shown in blue with belts shown in red.}
	\label{fig:leg_design}
	\vspace{-1.5em}
\end{figure}

To increase dynamic performance further, several designs move the ankle
actuators above the knee and closer to the hip. 
The MIT Humanoid \cite{mit_humanoid} places both the knee
and ankle actuators on the thigh, driving each through a belt transmission.
The resulting knee-ankle coupling is linear, and the ankle has only pitch control.
The Tello Leg \cite{tello_leg} drives the knee and ankle jointly
through a differential pair of actuators mounted near the hip, again a
linear coupling between the two joints. 
Xiao et al. \cite{Xiao_knee_ankle_pkm} similarly move all
actuators above the knee using a parallelogram mechanism in the thigh
combined with a 2-RSS PKM, and derive the resulting
knee-ankle coupling explicitly.
PAL Kangaroo \cite{MINGOHOFFMAN2024104827} places all its leg actuators,
including the ankle's, near the hip using linear actuators, resulting in a
nonlinear serial-parallel transmission.

In this work, we present a coupled series-parallel transmission leg (Fig.~\ref{fig:leg_design}) 
following the same design objective of proximal actuator
placement. The knee joint itself is realized as a PKM with a double
four-bar linkage, giving it a nonlinear transmission ratio. The ankle is
likewise a PKM of type 2RSU+1U, actuated in both pitch and roll. 
The design places both ankle actuators closer to the hip that
are routed through gears coaxial to the knee joint via
fixed belt reduction.
The novelty lies in the structural, nonlinear knee-ankle coupling as it
is a direct consequence of the transmission layout.
It lets the knee and
ankle share actuator effort dynamically, reducing peak torque and
velocity demand on each joint beyond what either joint could achieve
acting independently.
Unlike the linear belt coupling \cite{mit_humanoid} or constant
differential couplings used in prior proximal-actuation designs, ours require
configuration dependent nonlinear transmissions.
The presented design uses rotary actuators in contrast to linear actuators in the Kangaroo
leg design \cite{MINGOHOFFMAN2024104827}.

\paragraph*{Contribution}
First, we present a detailed kinematics analysis of the PKMs
used in the leg. We derive the nonlinear knee-ankle coupling
transmissions and study their effect on the achievable torque and
velocity envelope of each joint involved in the coupling. 
Secondly, our analysis provides an
in depth understanding of the nonlinear couplings more broadly between PKMs. 
Third, we quantify the resulting mechanical advantage through a squat simulation.
We separately verify the coupling on hardware through direct torque distribution 
measurements and PD (proportional-derivative) tracking of fast foot motions. 
Together, these show that the coupling lets the knee and ankle share actuator 
capabilities dynamically, 
beyond what either could achieve alone. The
leg also achieves a low reflected
mass at the foot when compared to similar designs.


\paragraph*{Organization}
The rest of this paper is organized as follows. Sec.~\ref{sec:preliminaries}
introduces the mathematical notation used throughout the paper.
Sec.~\ref{sec:design} describes the leg's design.
Sec.~\ref{sec:kinematics} presents the kinematic analysis and analyzes the
nonlinear coupling. Sec.~\ref{sec:results} presents the
simulation and hardware verification results. 
Sec.~\ref{sec:conclusion} concludes the paper and discusses future work.

\section{Preliminaries}
\label{sec:preliminaries}
\subsection{Loop Closure Constraints}
Let $(\mathbf{q}, \dot{\mathbf{q}}, \ddot{\mathbf{q}})$ represent the $n$-dimensional position, velocity and acceleration vectors of the joints in the spanning tree of the robot.
The vectors $(\mathbf{y}, \dot{\mathbf{y}}, \ddot{\mathbf{y}})$ represent the $m$-dimensional independent position, velocity and acceleration variables for a regular configuration. 
Loop constraints are algebraic constraints that must be satisfied by the motion of a multi-body system. The constraints can be expressed in implicit
$\bigl(\boldsymbol{\phi}(\mathbf{q})=\mathbf{0}\in \mathbb{R}^{n_c} \bigr)$ and explicit
$\bigl(\mathbf{q}=\boldsymbol{\gamma}(\mathbf{y}) \in \mathbb{R}^n \bigr)$ forms, 
summarized in Table~\ref{tab:loop_constraints}.
\begin{table}[h]
	\centering
	\small
	\caption{Loop closure constraints~\cite{featherstone2008rigid}}
	\label{tab:loop_constraints}
	\begin{tabular}{@{}lccc@{}}
		\toprule
		Type & Position & Velocity & Acceleration \\
		\midrule
		Implicit & $\bm{\phi}(\mathbf{q}) = \mathbf{0}$
		& $\mathbf{K}\dot{\mathbf{q}} = \mathbf{0}$
		& $\mathbf{K}\ddot{\mathbf{q}} = \mathbf{k}$ \\
		Explicit & $\mathbf{q} = \bm{\gamma}(\mathbf{y})$
		& $\dot{\mathbf{q}} = \mathbf{G}\dot{\mathbf{y}}$
		& $\ddot{\mathbf{q}} = \mathbf{G}\ddot{\mathbf{y}} + \mathbf{g}$ \\
		\bottomrule
	\end{tabular}
	\vspace{-0.5em}
\end{table}
Here, $\mathbf{K}=\dfrac{\partial \bm{\phi}}{\partial \mathbf{q}} \in \mathbb{R}^{n_c \times n}$,
$\mathbf{k}=-\mathbf{K}\dot{\mathbf{q}}\in \mathbb{R}^{n_c}$,
$\mathbf{G}=\dfrac{\partial \bm{\gamma}}{\partial \mathbf{y}}\in \mathbb{R}^{n \times m}$,
and $\mathbf{g}=\dot{\mathbf{G}}\dot{\mathbf{y}}\in \mathbb{R}^{n}$.
Since $\bm{\phi}$ and $\bm{\gamma}$ express the same constraints,
the following can be deduced:
$\boldsymbol{\phi}\circ\boldsymbol{\gamma}=\mathbf{0}$,
$\mathbf{K}\mathbf{G}=\mathbf{0}$, and
$\mathbf{K}\mathbf{g}=\mathbf{k}$.
This paper follows the explicit form of equations which are derived numerically
from the implicit constraints via the Generalized Coordinate Partitioning method ~\cite{wehage1982generalized}. $\mathbf{G}$ and $\mathbf{g}$ are
obtained from the implicit constraints  as
\begin{equation}
	\mathbf G = \begin{bmatrix} \mathbf I \\ -\mathbf{K}_d^{-1} \mathbf{K}_i \end{bmatrix}, \quad
	\mathbf{g} = \begin{bmatrix} \mathbf{0} \\ \mathbf{K}_d^{-1} \mathbf{k} \end{bmatrix}. \\
	\label{eq:numerical_Gg}
\end{equation}

Let $\mathbf{u} \in \mathbb{R}^{p}$ denote the $p$ actuator coordinates, related
to the independent coordinates by
\begin{equation}
	\dot{\mathbf{u}} = \mathbf{G}_u(\mathbf{y})\, \dot{\mathbf{y}},
	\qquad \mathbf{G}_u \in \mathbb{R}^{p \times m}.
	\label{eq:actuator_jacobian}
\end{equation}
If the actuators coincide with independent joints, $\mathbf{G}_u$ can be extracted from $\mathbf{G}$ by selecting the rows
corresponding to the actuated joints. Using Jourdan's principle of virtual power,
it can be shown that 
\begin{equation}
	\mathbf{\tau}_y = \mathbf{G}^T \mathbf{\tau}
\end{equation}
where $\mathbf{\tau} \in \mathbb{R}^{n}$ is the vector of generalized forces of the spanning tree joints.
The torque at the actuators can be determined similarly, provided $\mathbf{G}_u$ is invertible
(see~\cite{kumar2020analytical}).
\begin{equation}
	\boldsymbol{\tau}_y = \mathbf{G}_u^T \boldsymbol{\tau}_u
	\quad\Longleftrightarrow\quad
	\boldsymbol{\tau}_u = \mathbf{G}_u^{-T} \boldsymbol{\tau}_y.
	\label{eq:tau_y_tau_u}
\end{equation}
\subsection{Reflected Mass Model}
By taking the first derivative of the forward kinematics, 
the Cartesian velocity at the foot is obtained as
\begin{equation}
	\dot{\mathbf{x}}  = \mathbf{J} (\mathbf{q}) \dot{\mathbf{q}} = 
	\begin{bmatrix}
		\mathbf{J}_v (\mathbf{q}) \\
		\mathbf{J}_w (\mathbf{q})
	\end{bmatrix} \dot{\mathbf{q}}
	\label{eq:eom_explicit}
\end{equation}
where $\mathbf{J} \in \mathbb{R}^{6 \times n}$ is the foot Jacobian matrix, and $\mathbf{J}_v \in \mathbb{R}^{3 \times n}$,
and $\mathbf{J}_w \in \mathbb{R}^{3 \times n}$ represent the linear and angular part of the Jacobian matrix respectively.
The Operational space inertia matrix is given by
\begin{equation}
	\bm{\Lambda}(\mathbf{q})
	=
	\left(
	\mathbf{J}
	\mathbf{M}^{-1}
	\mathbf{J}^{T}
	\right)^{-1}
\end{equation} 
where $\mathbf{M} \in \mathbb{R}^{n \times n}$ is the generalized mass inertia matrix.
\begin{equation}
	\bm{\Lambda}(\mathbf{q})^{-1}
	=
	\mathbf{J}(\mathbf{q})
	\mathbf{M}(\mathbf{q})^{-1}
	\mathbf{J}(\mathbf{q})^{T}
	=
	\begin{bmatrix}
		\bm{\Lambda}_{v}(\mathbf{q})^{-1}
		&
		\overline{\bm{\Lambda}}_{vw}(\mathbf{q})
		\\[4pt]
		\overline{\bm{\Lambda}}_{vw}(\mathbf{q})^{T}
		&
		\bm{\Lambda}_{w}(\mathbf{q})^{-1}
	\end{bmatrix},
\end{equation}
where
$\bm{\Lambda}_{v}(\mathbf{q})^{-1}$,
$\bm{\Lambda}_{w}(\mathbf{q})^{-1}$,
and
$\overline{\bm{\Lambda}}_{vw}(\mathbf{q})^{T}
\in \mathbb{R}^{3\times3}$
are the translational, rotational, and coupling matrices,
respectively.
The reflected mass at the foot quantifies the impact of the foot when colliding with an external
environment~\cite{mrm} and is given by
\begin{equation}
	m_{\hat n}(\mathbf{q})
	=
	 \left[\hat{\mathbf{n}}^T \bm{\Lambda}_v(\mathbf{q})^{-1} \hat{\mathbf{n}}\right]^{-1}
	\label{eq:reflected_mass}
\end{equation}
where $\hat{\mathbf{n}} \in \mathbb{R}^3$ is the evaluation direction. A lower reflected mass allows faster foot acceleration and gentler impacts, giving the leg greater force-control bandwidth during stance.

\section{Coupled Series - Parallel Hybrid Leg Architecture}
\label{sec:design}
\subsection{Design Details}
The coupled series-parallel transmission leg can be seen in Fig. \ref{fig:leg_design}, which 
is equipped with $6$ actuated degrees of freedom (DOF). It is
designed to be lightweight and weighs approximately \qty{6.1}{\kilogram}.
\begin{figure}[!h]
	\centering
	\begin{minipage}{0.48\linewidth}
		\centering
		\includegraphics[width=\linewidth]{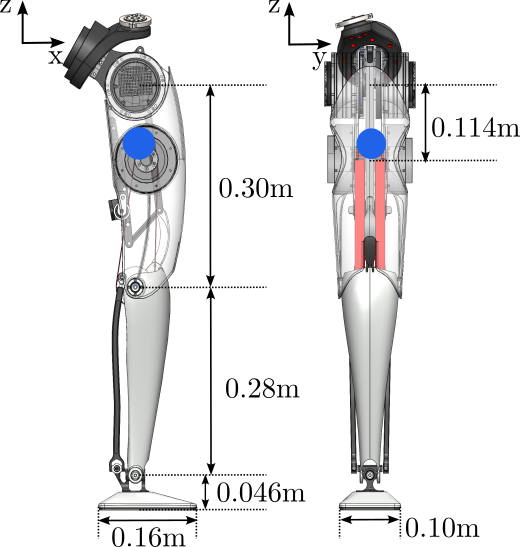}
	\end{minipage}%
	\hspace{0.2em}
	\begin{minipage}{0.40\linewidth}
		\centering
		\small
		\begin{tabular}{@{}lr@{}}
			\toprule
			Link & Mass (kg) \\
			\midrule
			Thigh  & 4.073 \\
			Shank   & 0.703 \\
			Ankle \& Foot & 0.376 \\
			\bottomrule
		\end{tabular}
	\end{minipage}
	\caption{Important dimensions for the leg (blue denotes the center of mass location), with the mass distribution of the leg.}
	\label{fig:leg_design_dimension}
	\vspace{-0.3em}
\end{figure}
Fig.~\ref{fig:leg_design_dimension} details the corresponding link lengths and mass distribution where the thigh spans \qty{0.3}{\meter} and carries the largest mass of around \qty{4.07}{\kilogram}, the shank spans \qty{0.28}{\meter} and carries \qty{0.70}{\kilogram}, and the ankle and foot together add only \qty{0.30}{\kilogram}.
The overall center of mass of the leg lies only \qty{0.114}{\metre} below the hip pitch axis and is denoted in blue.
The hip yaw, roll, and pitch actuators  are driven directly at the hip, and knee actuator is placed co-axially to the hip pitch actuator. The ankle actuators are placed closer to the hip, just below the hip and knee actuator, concentrating
the subsequent mass of the leg here. The ankle actuators' drive crosses the knee joint before reaching the ankle mechanism, which is described as the knee-ankle coupling later in Sec.~\ref{sec:kinematics:coupling}. This proximal placement keeps the shank and foot lightweight and this reduces the leg's inertia. Two commercially available actuators are used in the leg. The hip yaw, roll and ankle joints are actuated by 
Cubemars AK80-8 actuators~\cite{cubemars_ak_manual}, whereas the hip pitch and knee joints are actuated by
Cubemars AK10-9 actuators~\cite{cubemars_ak_manual}. 

\begin{figure}[!h]
	\centering
	\includegraphics[width=0.65\linewidth]{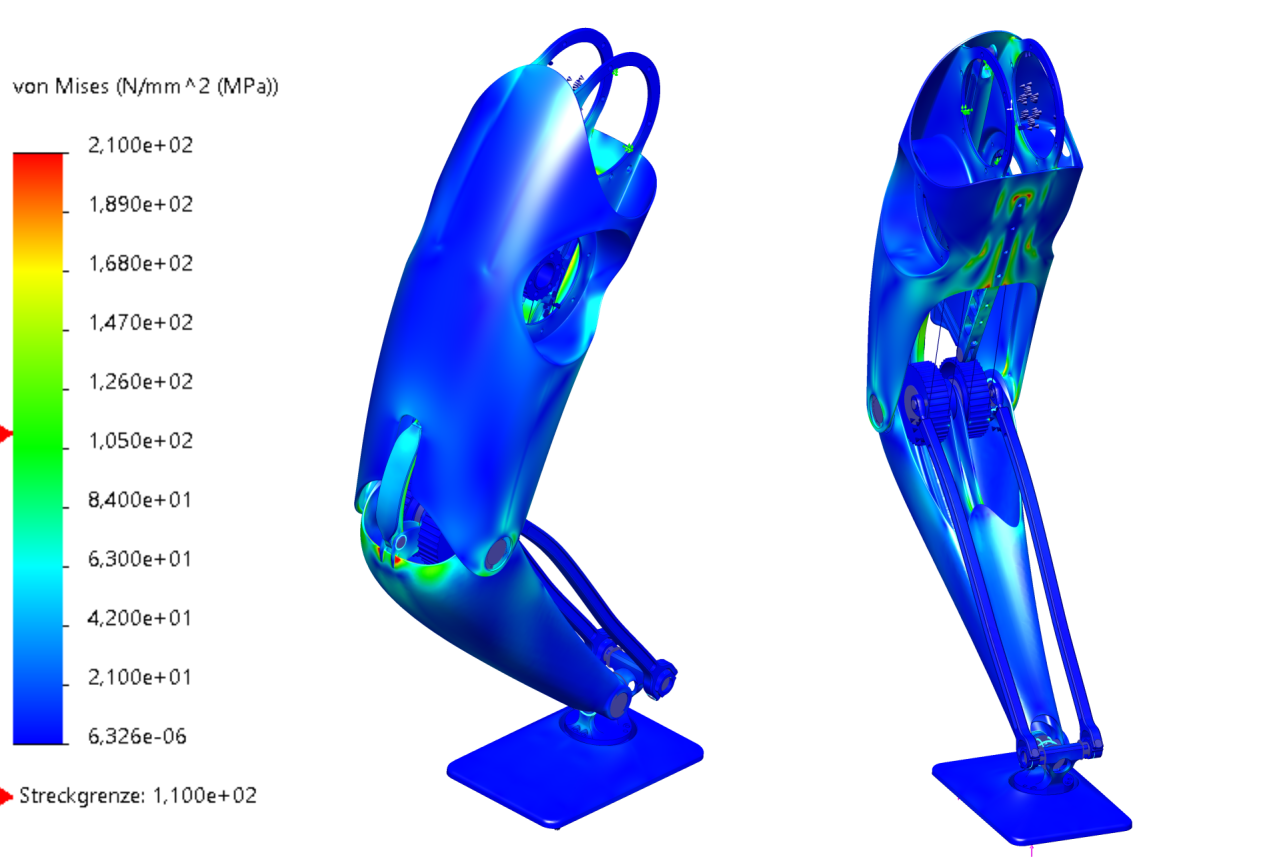}
	\caption{Finite element analysis with 600 N vertical force at the foot.}
	\label{fig:fea_analysis}
	\vspace{-1.0em}
\end{figure}
When designing a leg with lower distal inertia, careful selection of the structural components is unavoidable. Distributing the mass in the thigh, leaves the shank and foot more slender and therefore increases their possibility for elastic deflection under load. To keep the leg lightweight while maintaining stiffness, aluminum castings (AlSi10Mg) with stiff outer shell structures are used.   
Fig. \ref{fig:fea_analysis} shows the \textit{von Mises stress} remaining well below the alloy's yield strength of \qty{230}{\mega\pascal} across most of the structure. 
Similar scale humanoids weigh around \qtyrange{25}{30}{\kilogram} where each leg is expected
to support the full weight of the robot during dynamic maneuvers .
The maximum force of \qty{600}{\newton} is chosen with a
safety factor of \num{2.5}, where failures are possible on bigger impacts.
Fig.~\ref{fig:fea_analysis} shows the weaker points of the design at the thigh's backside and closer to the knee joint attachment. These areas are planned to have more reinforcements in cases of hardware failures.


\subsection{Parallel Submechanisms}
\begin{figure}[t]
	\centering
	\includegraphics[width=0.35\linewidth]{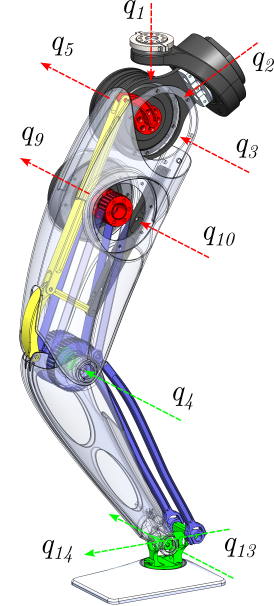}
	\includegraphics[width=0.40\linewidth]{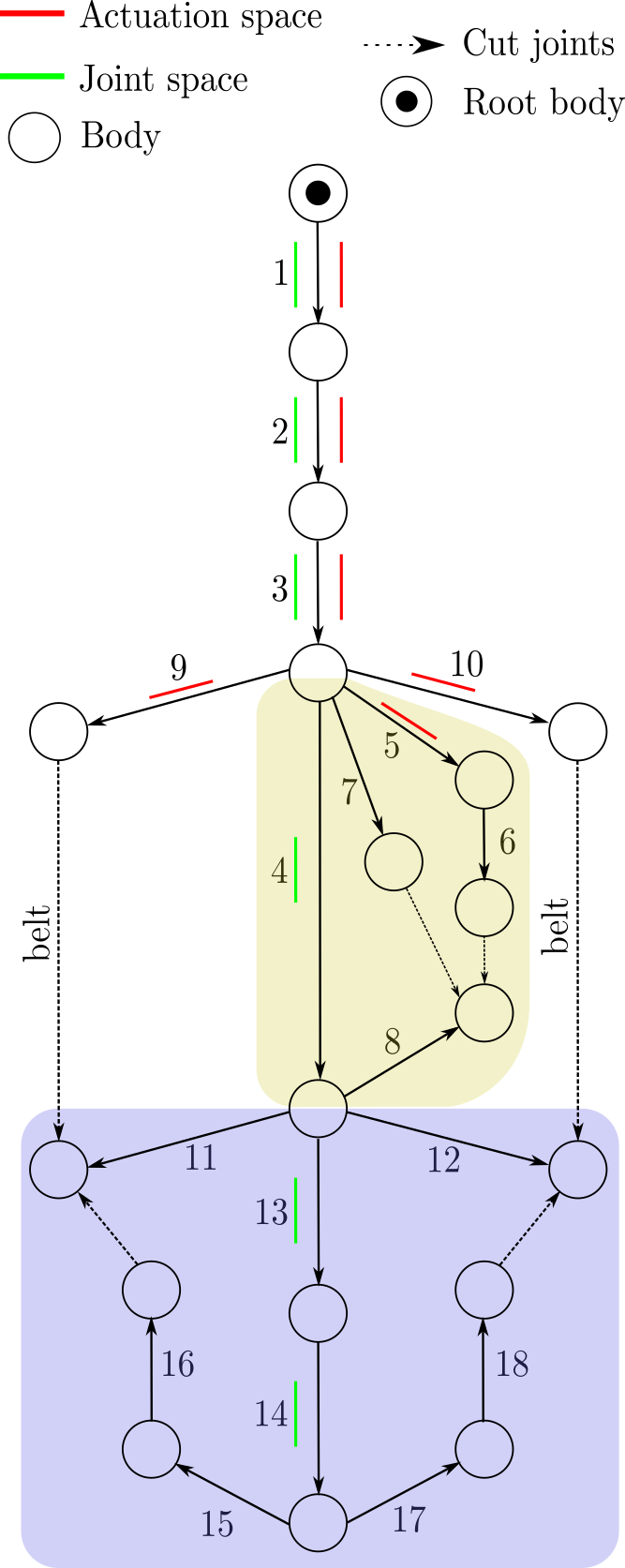}
	\caption{Parallel mechanism (left) and topological graph (right) of the leg.}
	\label{fig:topological_graph}
	\vspace{-1.5em}
\end{figure}
Fig. \ref{fig:topological_graph} summarizes the parallel mechanisms in the leg. In the
figure, actuators are denoted in red and independent joints are
denoted in green. Cut joints are denoted by dotted lines. The leg is a coupled
series-parallel hybrid kinematic chain formed by three submechanisms:
\begin{itemize}
    \item 3R Hip Yaw, Roll, and Pitch (3 DOF)
    \item Planar closed-loop Knee (1 DOF, yellow in Fig.~\ref{fig:topological_graph})
    \item 2\{RSU\} + 1U Ankle Pitch/Roll (2 DOF, blue in Fig.~\ref{fig:topological_graph}).
\end{itemize}

The hip is a conventional open (serial) chain, all three actuators are directly and
independently driven at their respective joints. These joints are denoted as $q_1$,
$q_2$, and $q_3$ for hip yaw, roll and pitch respectively, in
Fig.~\ref{fig:topological_graph}.
 
The knee is a single-DOF closed-loop linkage, one motor-driven crank (joint $q_5$) in
the thigh closes through two rocker links onto a shared coupler joint. One rocker link
is grounded at the hip-pitch link and the other is grounded at the knee output link.
The independent joint for knee flexion and extension is denoted as $q_4$.

The ankle is a 2-DOF closed-loop differential mechanism, a central pitch/roll universal
joint between the knee and the ankle output plate. Two gears mounted along the
knee joint axis, denoted as $q_{11}$ and $q_{12}$,  has a short output lever
with a spherical joint at its tip. A rod runs from that spherical joint to the
universal joint, right behind the foot plate. The two rods together drive the 
universal joint and thus the ankle pitch and roll differentially. The ankle dorsiflexion and
plantar flexion joint is denoted as $q_{13}$ and ankle inversion and eversion (roll) is
denoted as $q_{14}$.

The two ankle actuators, $q_9$ and $q_{10}$, drive the gears, $q_{11}$ and $q_{12}$,
coaxial with the knee joint, via timing belts running through the thigh, at a constant
1.6:1 reduction. This couples the two submechanisms and the next section discusses how this 
nonlinear coupling affects the leg.

\section{Kinematic Analysis of Coupled Parallel Submechanisms}
\label{sec:kinematics} 



\subsection{Knee Parallel Mechanism}
\begin{figure}[!h]
	\centering
	\resizebox{0.535\linewidth}{!}{%
		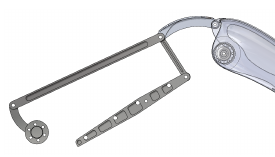
	}
	\includegraphics[width=0.45\linewidth]{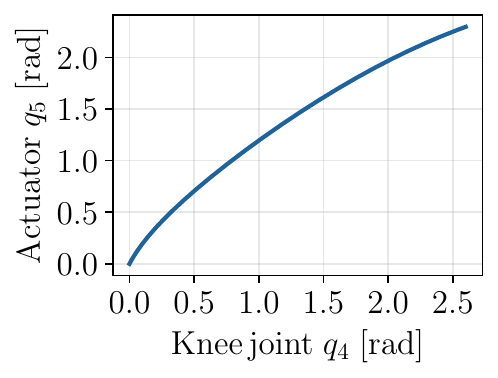}
	\caption{Knee submechanism with joint notations (left) and configuration space (right).}
	\label{fig:knee_mechanisms_annotated}
	\vspace{-0.5em}
\end{figure}

The knee submechanism can be seen as a PKM in
Fig.~\ref{fig:topological_graph} marked in light yellow.
The submechanism has a total of 5 spanning tree joints where the independent joint is
$\mathbf{y}_{knee} = q_4$,
and the actuator joint is $\mathbf{u}_{knee} = q_5$. Due to two four bar mechanisms, it imposes two cut 
joints between $l_6$, $l_7$ and $l_8$(Fig.~\ref{fig:knee_mechanisms_annotated} left). 
Thus, the submechanism imposes four loop closure constraints (2 translation constraints per cut joint).

The links $l_a$ and $l_b$ and the angle $\mu$ between them are fixed. 
For deriving the analytical map from the knee joint to motor angle, 
the four bar at the knee can be split into two triangles with the diagonal $l_{4b}$.
The angle each triangle contributes at that diagonal sums to $q_7$:
\begin{align}
l_{4b} &= \sqrt{l_4^2 + l_b^2 - 2 l_4 l_b \cos q_4'} \\
q_{7,1} &= \arcsin\!\left(\frac{l_4}{l_{4b}} \sin q_4'\right) \\
q_{7,2} &= \arccos\!\left(\frac{l_8^2 - l_7^2 - l_{4b}^2}{-2 l_7 l_{4b}}\right) \\
q_7 &= q_{7,1} + q_{7,2}
\end{align}
Analogously, the diagonal $l_{a7}$ splits the second four-bar into two triangles, and
$q_5$ is assembled from the angle each triangle contributes at that diagonal:
\begin{align}
l_{a7} &= \sqrt{l_7^2 + l_a^2 - 2 l_a l_7 \cos (\mu - q_7)} \\
q_{5,1} &= \arcsin\!\left(\frac{l_7}{l_{a7}} \sin (\mu - q_7)\right) \\
q_{5,2} &= \arccos\!\left(\frac{l_6^2 - l_5^2 - l_{a7}^2}{-2 l_5 l_{a7}}\right)
\end{align}
\begin{align}
	\quad q_5 = -\Big(q_{5,1} + q_{5,2}\Big)
	\label{eq:knee_pos}
\end{align}
The analytical map between the joint and the actuator (as in Eq.~\ref{eq:knee_pos}) can
be seen in Fig.~\ref{fig:knee_mechanisms_annotated} (right).
Position constraints in Eq.~\ref{eq:knee_pos} are differentiated once to obtain the velocity level constraints.
\begin{equation}
	\dot{\mathbf{q}}_{knee} = \mathbf{G}_{knee}\, \dot{\mathbf y}_{knee},
	\qquad \mathbf{G}_{knee} \in \mathbb{R}^{5 \times 1}.
	\label{eq:knee_vel}
\end{equation}

The actuator row $\mathbf{G}_{u,\mathrm{knee}}$, corresponding to $q_5$, is extracted
from $\mathbf{G}_{knee}$, giving the nonlinear knee transmission ratio $G_{u,\mathrm{knee}}$ such that,
\begin{equation}
	\dot{q}_5 = G_{u,\mathrm{knee}}\, \dot{q}_4.
	\label{eq:knee_transmission_ratio}
\end{equation}
\begin{figure}[t]
    \centering
    \includegraphics[width=0.49\linewidth]{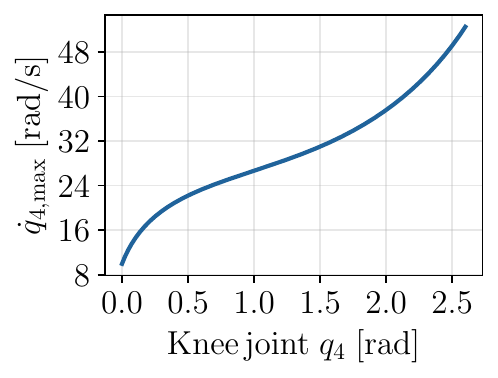}
    \includegraphics[width=0.49\linewidth]{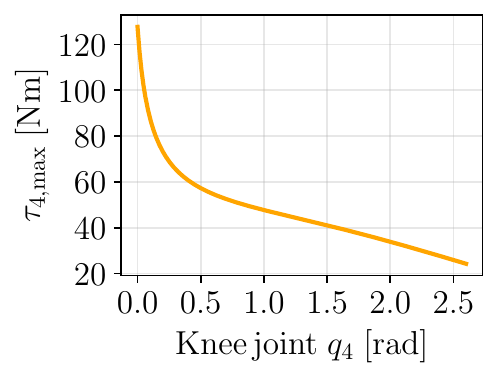}
    \caption{Maximum knee velocity (left) and torque (right) for different configurations.}
    \label{fig:knee_transmission}
	\vspace{-1.5em}
\end{figure}
Fig.~\ref{fig:knee_transmission} shows the resulting envelope where the actuator's rated
velocity \qty{24}{\radian\per\second} and torque \qty{53}{\newton\metre} are scaled by the configuration 
dependent nonlinear transmission ratio using Eq.~\ref{eq:actuator_jacobian} and ~\ref{eq:tau_y_tau_u}. 
The transmission ratio reaches its maximum
$G_{u,\mathrm{knee}} = 2.41$ at the straight knee ($q_4 = 0$) where most of the
support is required by the leg from the knee. It falls to
its minimum of $0.49$ at the fully bent knee 
($q_4 \approx \qty{2.6}{\radian}$) 
and the knee's maximum achievable velocity is the highest.
In our use case, near full extension, the
knee bears the body's weight during stance and needs high torque. Near full flexion,
the leg is unloaded during swing and needs high speed instead to clear the foot
quickly.

\subsection{Ankle Parallel Mechanism}
The ankle's submechanism of type 2\{RSU\}+1U can also be seen as a PKM 
in Fig.~\ref{fig:topological_graph} marked in light blue.
The submechanism has a total of 8 spanning tree joints and the
chosen
output coordinates for this mechanism are $q_{11}$ and $q_{12}$ representing coaxial
gears at the knee location. 
This submechanism considers 2 spherical joints as cut joints and imposes 6 constraints
(3 translation constraints per cut joint).
In summary, $\mathbf{y}_{ankle}=[q_{13},q_{14}]^T$ and
$\mathbf{u}_{ankle}=[q_{11},q_{12}]^T$. 
\begin{figure}[!h]
	\centering
	\includegraphics[height=0.45\linewidth]{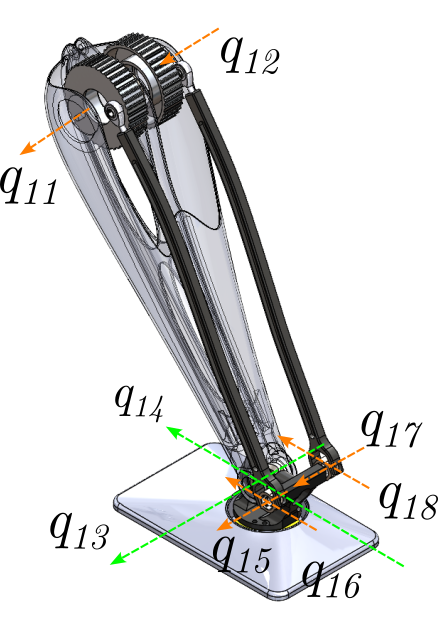}
	\includegraphics[width=0.45\linewidth]{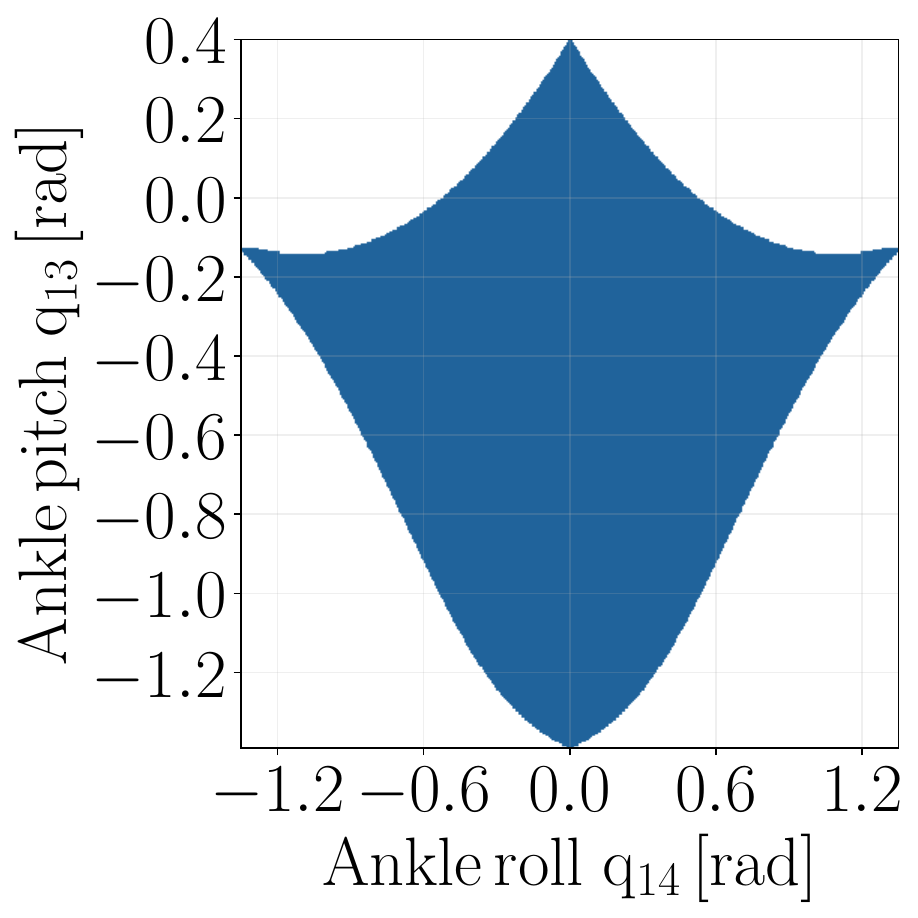}
	\caption{Ankle joints notation (left) and feasible configuration space (right).}
	\label{fig:ankle_cspace}
	\vspace{-0.5em}
\end{figure}

On position level, the loop closures are resolved numerically with iterative Newton-Raphson method.
The configuration space of the mechanism is shown in Fig.~\ref{fig:ankle_cspace}, where
only the feasible configurations are plotted within the limits imposed by 
$\qty{-1.68}{\radian} \leq q_{11},q_{12} \leq \qty{0.73}{\radian}$.
The velocity and acceleration level constraints can be obtained by following Eq.~\ref{eq:numerical_Gg}.
\begin{equation}
	\dot{\mathbf{q}}_{ankle} = \mathbf{G}_{ankle}\, \dot{\mathbf{y}}_{ankle},
	\qquad \mathbf{G}_{ankle} \in \mathbb{R}^{8 \times 2}.
	\label{eq:ankle_vel}
\end{equation}
The actuator rows $\mathbf{G}_{u,\mathrm{ankle}}$, corresponding to $q_{11}$ and
$q_{12}$, can be extracted from $\mathbf{G}_{ankle}$ as
\begin{equation}
	\dot{\mathbf{u}}_{ankle} = \mathbf{G}_{u,\mathrm{ankle}}\, \dot{\mathbf{y}}_{ankle},
	\qquad \mathbf{G}_{u,\mathrm{ankle}} \in \mathbb{R}^{2 \times 2}.
	\label{eq:ankle_transmission_ratio}
\end{equation}

\begin{figure}[!h]
    \centering
    \includegraphics[width=0.49\linewidth]{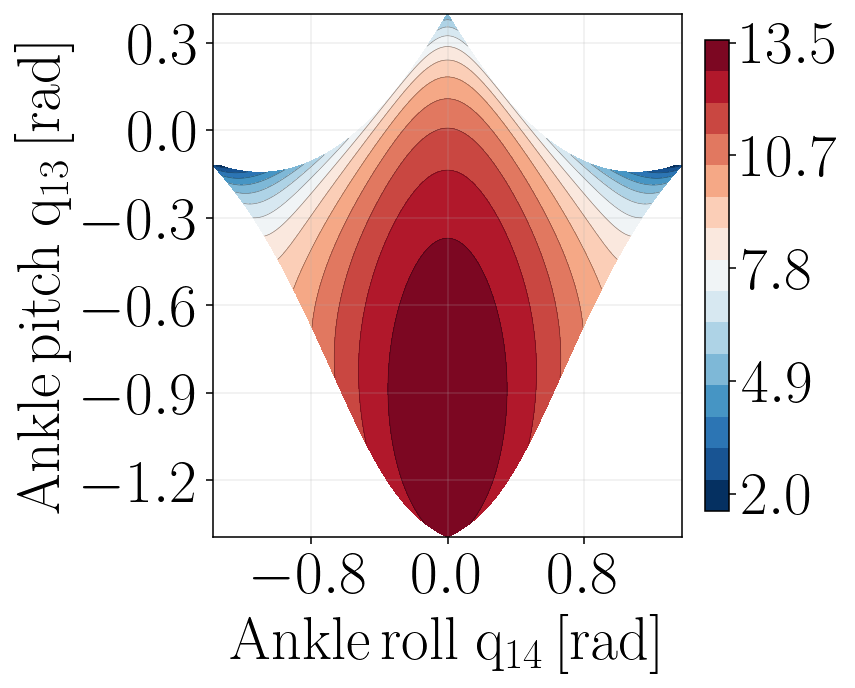}
    \includegraphics[width=0.49\linewidth]{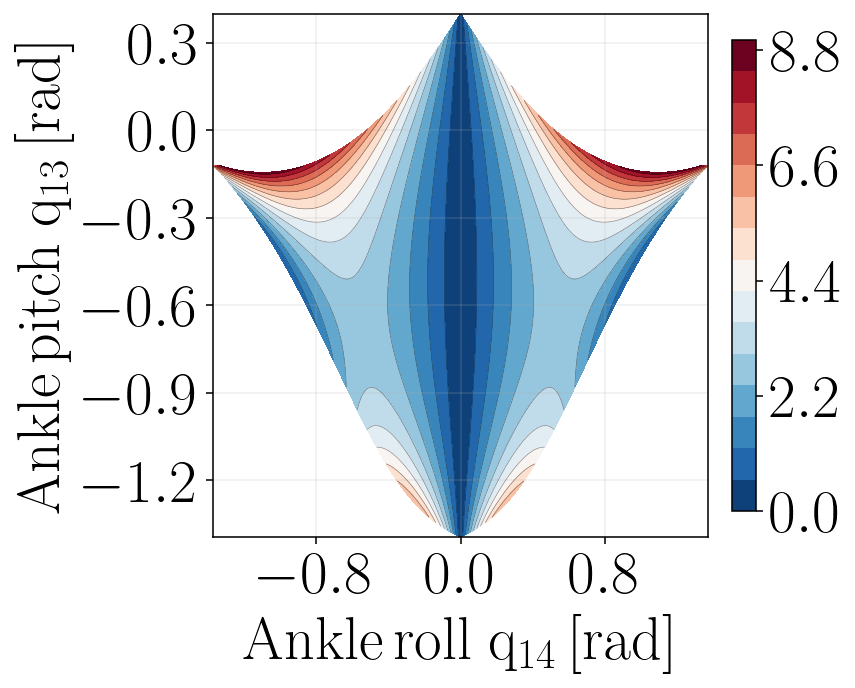}
    \caption{Maximum ankle pitch velocity (left) and roll velocity (right) across the ankle configuration space.}
    \label{fig:ankle_cspace_vel}
	\vspace{-1.0em}
\end{figure}
\begin{figure}[!h]
    \centering
    \includegraphics[width=0.49\linewidth]{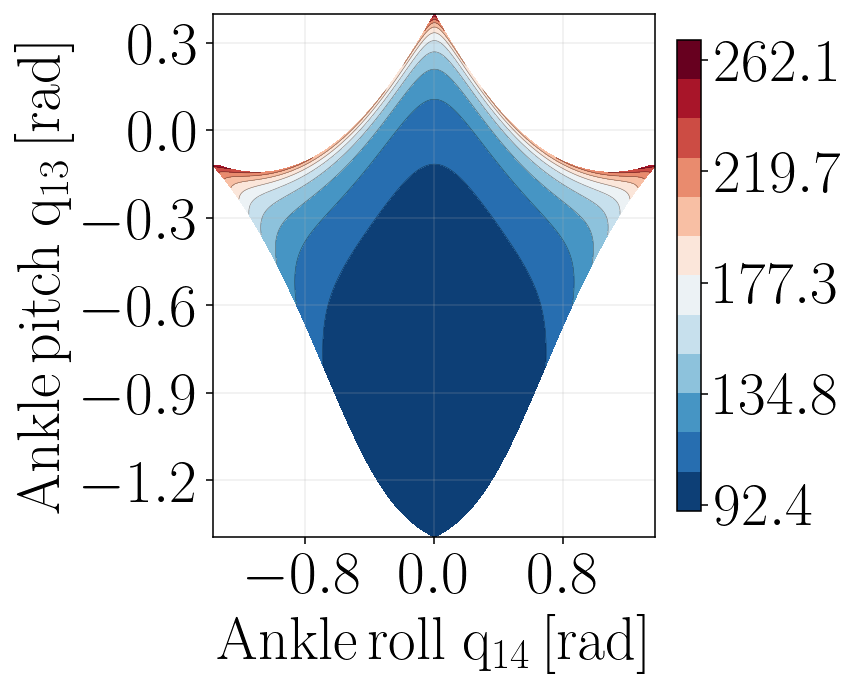}    
    \includegraphics[width=0.49\linewidth]{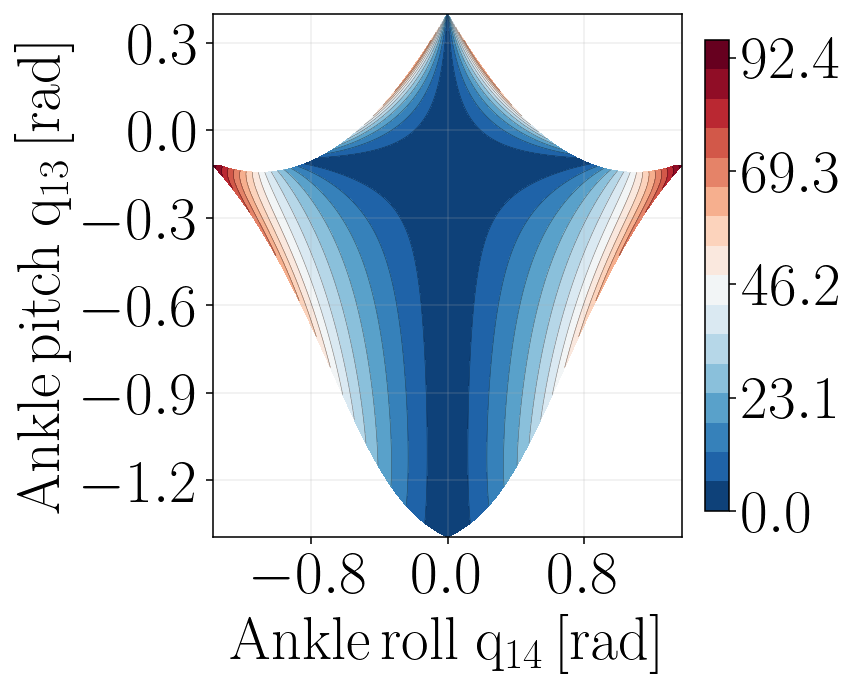}
    \caption{Maximum ankle pitch torque (left) and roll torque (right) across the ankle configuration space.}
    \label{fig:ankle_cspace_tau}
	\vspace{-0.5em}
\end{figure}
Fig.~\ref{fig:ankle_cspace_vel} shows the maximum ankle velocity.
The ankle actuators are rated \qty{15.625}{\radian\per\second} and \qty{40}{\newton\metre} at the
gear location. Following Eq.~\ref{eq:actuator_jacobian},
ankle pitch velocity ranges from \num{1.6} to \qty{14.4}{\radian\per\second} and ankle roll velocity reaches up to
\qty{9.05}{\radian\per\second}. Pitch movement is far more sensitive to ankle configuration than roll movement.
Fig.~\ref{fig:ankle_cspace_tau} shows the maximum ankle torque where
ankle pitch torque ranges from \num{90} to \qty{270}{\newton\metre} and ankle roll torque reaches up to
\qty{105}{\newton\metre}. Both exceed a single actuator's \qty{40}{\newton\metre}
rating at the gear, since the two ankle motors act together through the nonlinear
transmissions. 

\subsection{Coupling Between Knee and Ankle PKM}
\label{sec:kinematics:coupling}
\subsubsection{Position Level Constraints}
\begin{figure}[!h]
	\centering
	\includegraphics[height=0.45\linewidth]{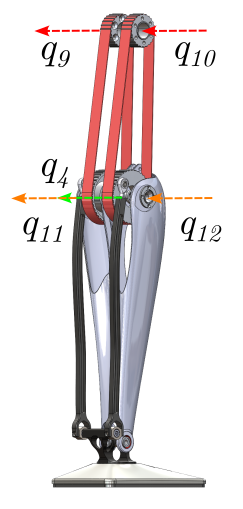}
	\includegraphics[width=0.50\linewidth]{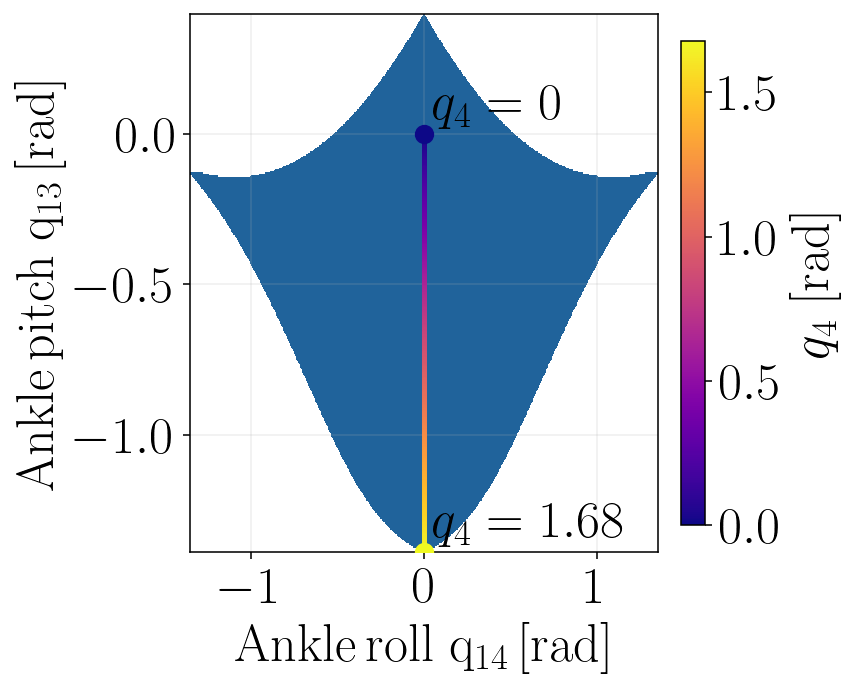}
	\caption{Coupling joints notation (left) and configuration space of the ankle pitch  with knee (right).}
	\label{fig:coupling_ankle_cspace}
	\vspace{-1.0em}
\end{figure}
The coupling between the knee and ankle is shown in
Fig.~\ref{fig:coupling_ankle_cspace}.
The ankle actuators denoted by joints $q_9$ and $q_{10}$, drive the gears,
$q_{11}$ and  $q_{12}$, through a fixed $1.6{:}1$ belt reduction that runs coaxially
through the knee joint. This adds a structural, position-level dependency on
the knee angle,
\begin{equation}
	q_{9} = 1.6(q_{11} + q_4), \qquad
	q_{10} = 1.6(q_{12} + q_4)
	\label{eq:coupling_position}
\end{equation}
where $q_{11}$ and $q_{12}$ can be obtained only from 
the ankle kinematics described before evaluated at $q_4 = 0$. The
knee and ankle submechanisms are no longer independent and hence the combined
independent coordinates are $\mathbf{y} = [q_4, q_{13}, q_{14}]^T$, and the
combined actuator coordinates are $\mathbf{u} = [q_{5}, q_{9}, q_{10}]^T$.

Fig.~\ref{fig:coupling_ankle_cspace} shows the position level consequence of
Eq.~\eqref{eq:coupling_position}.  With the ankle actuators held fixed, knee
flexion from \num{0} to \qty{1.68}{\radian} alone drags the ankle pitch angle
through part of its range.
At $q_4=\qty{1.68}{\radian}$, the ankle pitch reaches $\qty{-1.39}{\radian}$, purely from
the knee's motion.
\begin{figure*}[!t]
    \centering
    \subfloat[Visualization of the ankle pitch with knee movements (ankle motors held fixed).]{
        \includegraphics[width=0.7\linewidth]{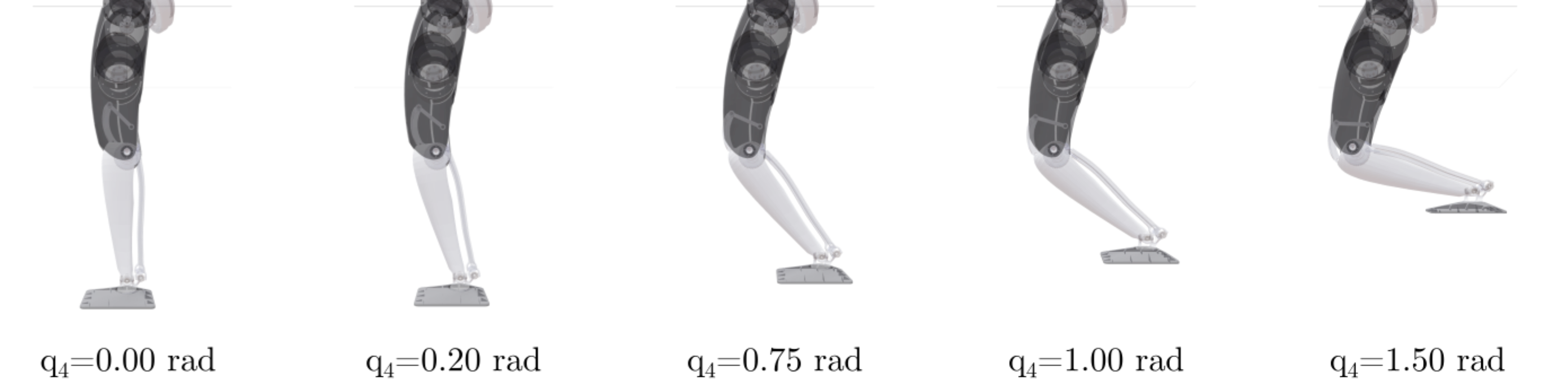}
    }
    \vspace{0.1em}
    \subfloat[Maximum ankle pitch velocity over the ankle configuration space for $\dot{q}_4=+\dot{q}_{4,\max}$. "+" marker denotes the current knee configuration.]{
        \includegraphics[width=0.9\linewidth]{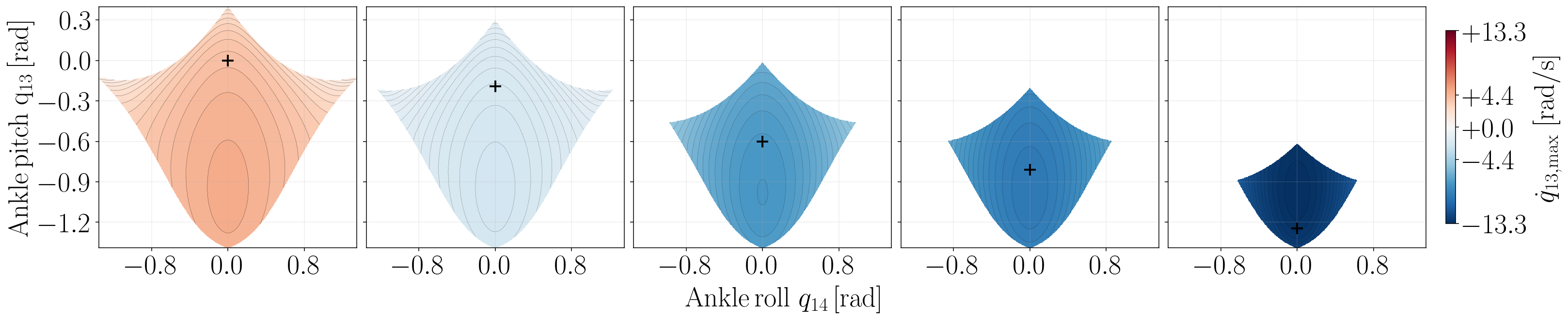}
    }
    \vspace{0.1em}
    \subfloat[Maximum ankle pitch velocity over the ankle configuration space for $\dot{q}_4=-\dot{q}_{4,\max}$. "+" marker denotes the current knee configuration.]{
        \includegraphics[width=0.9\linewidth]{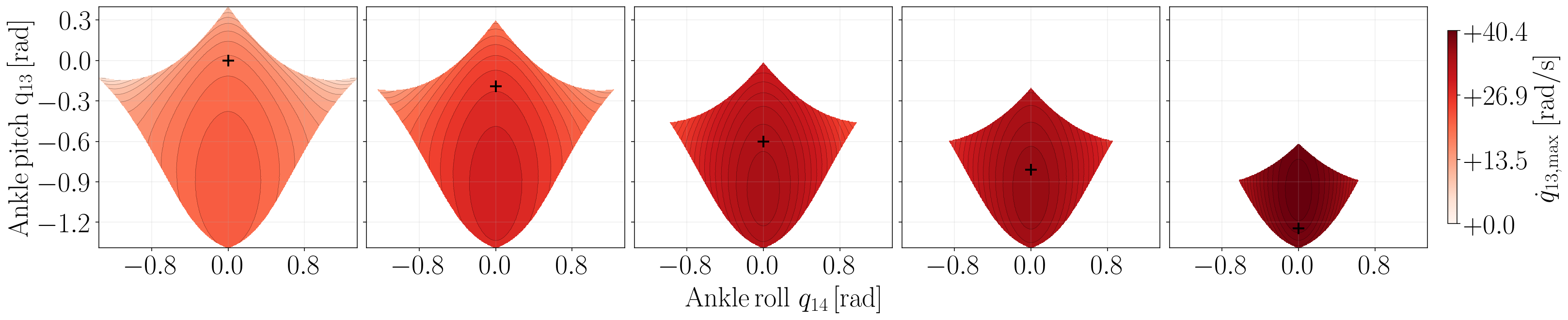}
    }
    \caption{Velocity-level coupling between the knee and ankle joints.}
    \label{fig:coupling_ankle_vel}
    \vspace{-1.5em}
\end{figure*}
\subsubsection{Velocity Level Constraints}
Differentiating Eq.~\eqref{eq:coupling_position} gives the combined transmission
Jacobian
\begin{equation}
	\mathbf{G}_u =
	\left[
	\begin{array}{ccc}
		G_{u,\mathrm{knee}} & 0 & 0 \\
		1.6 & \multicolumn{2}{c}{\multirow{2}{*}{$1.6\,\mathbf{G}_{u,\mathrm{ankle}}$}} \\
		1.6 & &
	\end{array}
	\right] \in \mathbb{R}^{3 \times 3},
	\label{eq:coupling_jacobian}
\end{equation}
The first column is exactly $[1.6, 1.6]^T$ for the ankle actuators, independent of configuration. 
With $\dot{\mathbf{u}} = \mathbf{G}_u\, \dot{\mathbf{y}}$, actuator velocities 
are given by
\begin{equation}
	\begin{bmatrix} \dot q_9 \\ \dot q_{10} \end{bmatrix}
	= 1.6\, \dot q_4 \begin{bmatrix}1\\1\end{bmatrix}
	+ 1.6\, \mathbf{G}_{u,\mathrm{ankle}} \begin{bmatrix} \dot q_{13} \\ \dot q_{14} \end{bmatrix}.
	\label{eq:coupling_velocity_ankle}
\end{equation}
This means that knee motion always acts together with ankle motion. 
To understand this coupling at the velocity level, Fig.~\ref{fig:coupling_ankle_vel}
visualizes the achievable ankle velocities at different knee configurations.
It assumes that ankle motors are fixed while moving the knee to particular configurations
$q_4 \in\{0.0,0.2,0.75,1.0,1.5\} \unit{\radian}$ in Fig.~\ref{fig:coupling_ankle_vel}(a). The maximum ankle
velocities are then computed based on maximum rated velocity of the motors.

Two edge cases are shown. 
In the first case in Fig.~\ref{fig:coupling_ankle_vel}(b), when the knee flexes ($+\dot q_{4,\max}$), the feasible
ankle pitch shifts toward dorsiflexion (negative pitch), and the maximum
achievable plantarflexion rate falls, eventually going negative. This means
the coupling drives the ankle toward dorsiflexion even if the ankle motors are
commanded to plantarflex at their own maximum. At $q_4 = \qty{0.2}{\radian}$, 
the blue region shows that achievable ankle
plantarflexion has already gone negative. The knee's maximum velocity
overrides the ankle motors entirely, rather than merely consuming their
budget.
In the second case in Fig.~\ref{fig:coupling_ankle_vel}(c), when the knee extends
($-\dot q_{4,\max}$), the achievable plantarflexion rate exceeds what the ankle
motors alone ($\dot y_\mathrm{ankle,max}$) could produce (compare $q_4=0$ and $q_4=1.5$ rad), and grows further the more flexed the knee started. 
In both cases, part of each ankle actuator's
rated velocity is committed to the knee's motion rather than the ankle's own
independent movement, so the overall ankle velocity workspace shrinks at a
given knee angle. Hence, the coupling redirects bandwidth rather than adding it for
free.
Such coupling is targeted for dynamic gaits where knee flexion assists ankle dorsiflexion, 
useful for foot clearance during swing, and knee extension assists ankle
plantarflexion, useful for push-off during stance.

\subsubsection{Torque Mapping}
On the torque mapping, Eq.~\ref{eq:coupling_jacobian} informs that
knee always receives the full ankle actuator torque with a fixed,
configuration independent constant. Applying Eq.~\ref{eq:tau_y_tau_u}
to Eq.~\ref{eq:coupling_jacobian} gives the knee's share of the coupled
torque,
\begin{equation}
	\tau_4 = G_{u,\mathrm{knee}}\, \tau_{5} + 1.6\tau_{9} + 1.6\tau_{10},
	\label{eq:coupling_torque}
\end{equation}
where $\tau_{5}$ is the knee's own actuator torque and $\tau_{9}, \tau_{10}$
are the ankle actuator torques.
With both ankle motors contributing their rated
torque in the same direction, it adds up to \qty{\pm 80}{\newton\metre}, extending the achievable knee torque from \qtyrange{127.7}{207.7}{\newton\metre}.
This benefit requires both ankle motors to act in the
same direction. If they instead oppose the knee, the same coupling can reduce
or reverse the knee's net torque instead, similar to velocity transmissions.
Note that ankle roll torques has no effect on the knee because 
of the differential component.


\section{Results and Discussion}
\label{sec:results} 
This section presents results validating the proposed coupled 
series-parallel transmission. We quantify the knee-ankle coupling's 
torque benefit through squat simulation. The coupling model is verified
on the hardware through torque distribution measurements, and  
tracking performance is evaluated under fast foot motions via PD control.
Lastly, we compare the leg's reflected mass to comparable designs in the literature. 
The results can also be found in the accompanying video.
%

\subsection{Squat Motion Simulation Verification}
To quantify the coupling's benefit, we compare the required knee motor torque 
$\tau_5$ with and without crediting the coupled ankle contribution across a squat motion
(i.e. Eq.~\ref{eq:coupling_torque}
versus the same equation with $\tau_9 = \tau_{10} =0$).
The squat motion is generated in Cartesian space where the foot is moved upward and forward from its
initial pose $[0.0,0.09,-0.7]^T$ to $[0.25,0.25,-0.4]^T$. 
A constant vertical force of \qty{150}{\newton} and a
moment of \qty{20}{\newton\metre} about the $y$-axis are applied at the foot,
approximating one leg's contribution to the robot's estimated 
\qty{30}{\kilogram} weight in double support.
\begin{figure}[!h]
	\centering
	\includegraphics[width=0.49\linewidth]{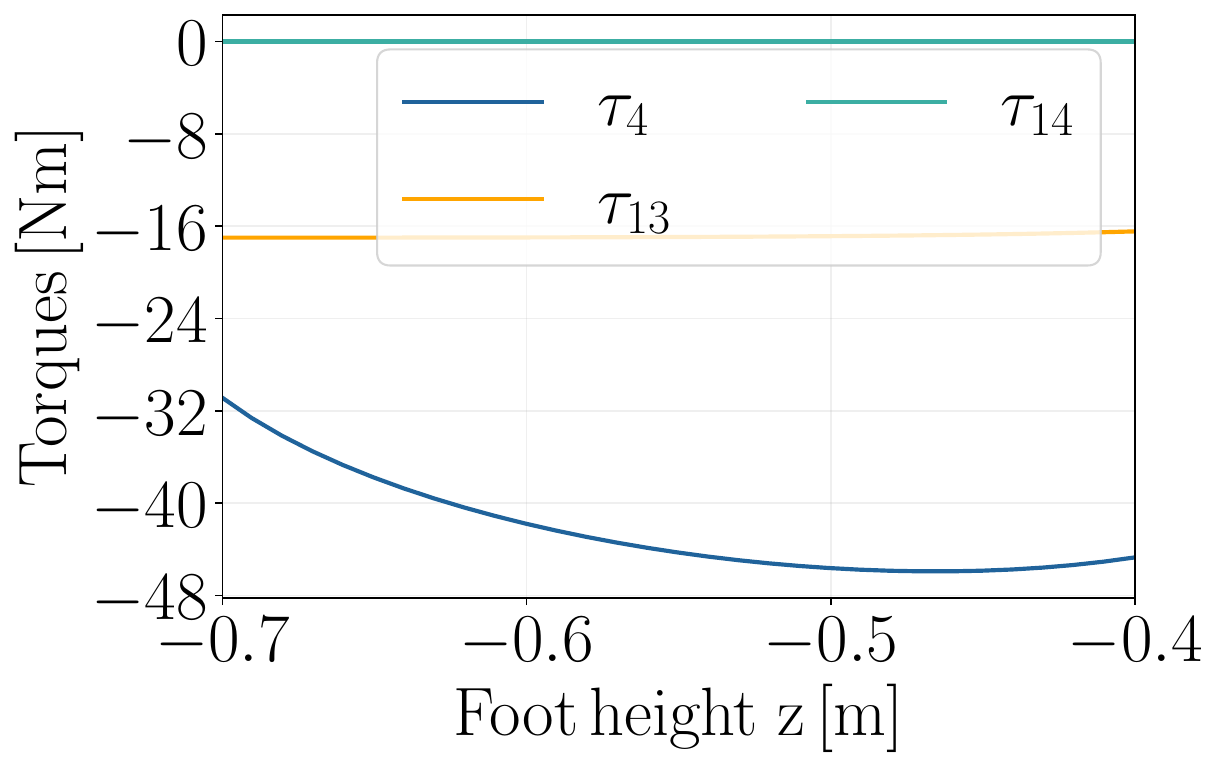}
	\hfill
	\includegraphics[width=0.49\linewidth]{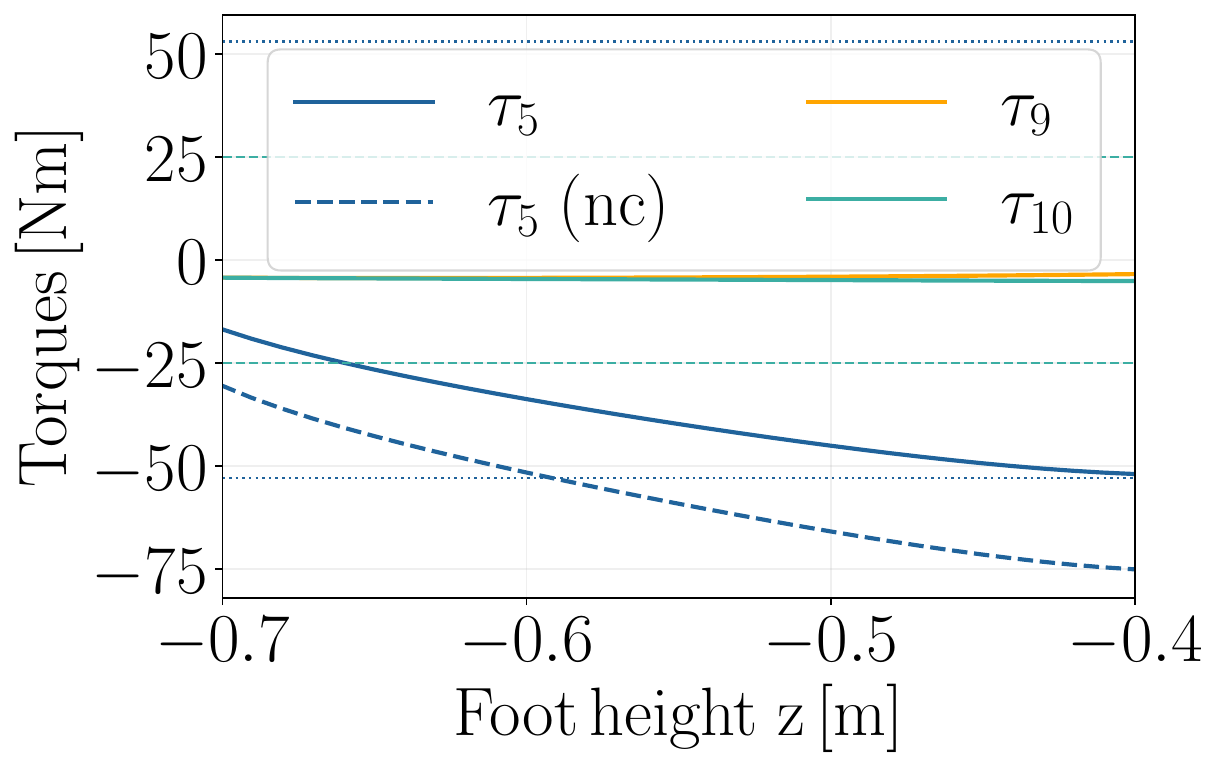}
	\caption{Independent joint torques (left) and motor torques (right) vs foot vertical position during the squat.}
	\label{fig:squat_tau}
	\vspace{-1.5em}
\end{figure}
Fig.~\ref{fig:squat_tau} (left) shows that the knee ($\tau_4$) and ankle pitch
($\tau_{13}$) carry the load, while ankle roll ($\tau_{14}$) stays near zero.
On the actuator side (right), the ankle actuators $\tau_9, \tau_{10}$ stay well 
within their $\pm25\,\mathrm{Nm}$ rating throughout the squat.

The peak demand in both cases occurs at the deepest squat pose at $z=-0.4\mathrm{m}$.
Without the coupling, the peak knee torque $\tau_5$ reaches $75~\mathrm{Nm}$ (denoted as
$\tau_5$(nc) in Fig.\ref{fig:squat_tau} right), 
which is $41.5\%$ above the $\pm 53~\mathrm{Nm}$ knee actuator rating. This makes 
 the trajectory infeasible for the knee motor alone. 
Broadly, the uncoupled torque exceeds rating for all positions beyond 
$\approx z=-0.6\mathrm{m}$. 
In contrast, with the coupling, the peak torque is reduced to $50.8~\mathrm{Nm}$, 
remaining within the actuator rating throughout the trajectory. 
Thus, the coupling decreases the peak knee torque demand by $32\%$ at the critical configuration. 
Because the nonlinear transmissions are configuration dependent and not
load dependent, required torque scales linearly with the applied foot load.
Scaling the \SI{150}{\newton}/\SI{20}{\newton\metre} load proportionally, the
maximum sustainable load before $\tau_5$ saturates rises from
$\SI{108}{\newton}/\SI{14.4}{\newton\metre}$ (uncoupled) to
$\SI{156}{\newton}/\SI{20.8}{\newton\metre}$ (coupled), a
$\SI{44.5}{\percent}$ increase in payload capacity.
Note that, in absence of ankle pitch torques similar to a linear squat, 
all the load is acting on the knee joint. In this case, the leg cannot squat
beyond $\approx z=-0.5\mathrm{m}$ with 150 N vertical force.

\subsection{Real System Nonlinear Coupling Verification}
To verify the knee-ankle coupling on the physical leg, a \qty{2.5}{\kilogram}
weight is attached \qty{0.5}{\metre} from the ankle, creating a constant
moment of $\qty{2.5}{\kg} \times \qty{0.5}{\m} \times\qty{9.81}{\m\per\s^2} \approx \qty{12.26}{\N\m}$ about the
ankle pitch axis. The knee and ankle actuators are commanded in pure feedforward
torque and the resulting joint positions and torques are logged on the real
system. With no position control active, $q_4$ and $q_{13}$
move freely in response to the applied and coupling torques under gravity.
\begin{figure}[!h]
	\centering
	\includegraphics[width=0.49\linewidth]{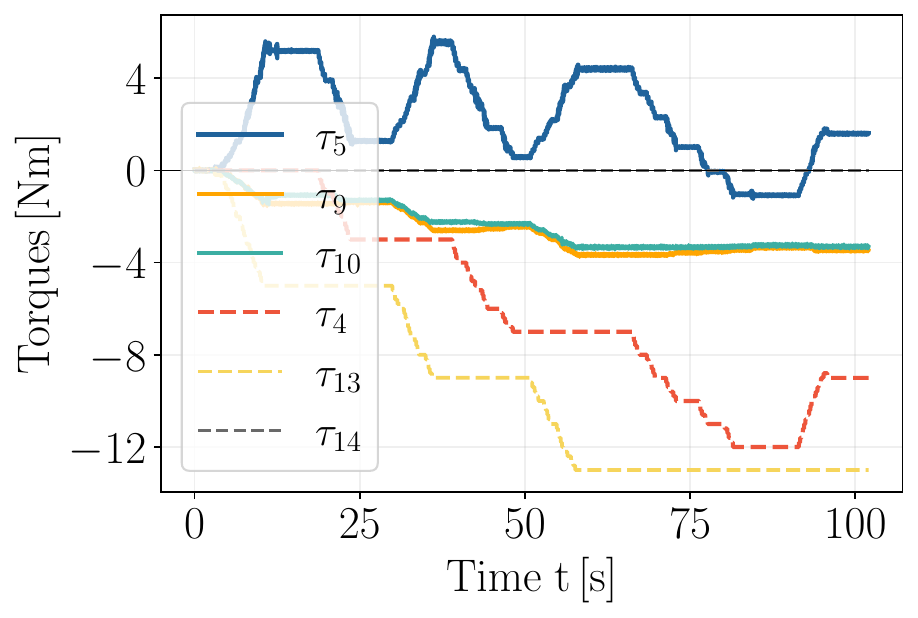}
	\includegraphics[width=0.49\linewidth]{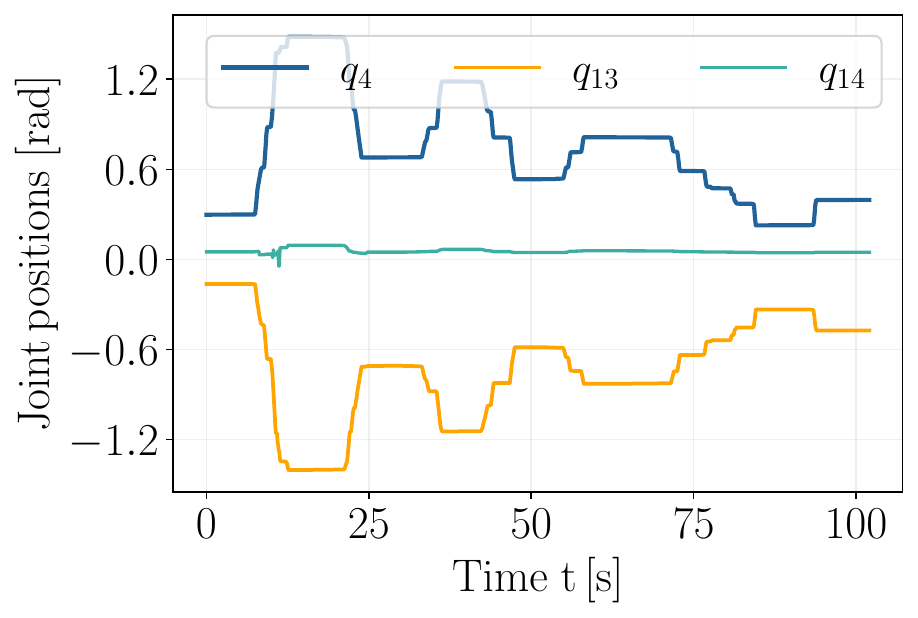}
	\caption{Coupling verification with an external load of 2.5kg at 0.5m distance.}
	\label{fig:hw_coupling}
	\vspace{-0.5em}
\end{figure}
Fig.~\ref{fig:hw_coupling} (left) shows the resulting independent joint
torques (dashed) and actuator torques (solid). 
At the start of the trace,
$\tau_4$ is held at zero, yet $\tau_5$ is already increasing. This
is the coupling torque from the ankle being injected directly into the
knee motor even though the knee itself is not commanded.
The system acts as if it was in serial configuration.
$\tau_4$ is then commanded to counteract this disturbance and 
hold the initial configuration.
As $\tau_4$ increases, $\tau_5$ drops
correspondingly, since coupled ankle
contribution now supplies part of the knee's required torque. 
This continues until $\tau_{13}$ settles near the theoretical holding value of
\SI{-12.26}{\newton\metre}, at which point $\tau_4$ is commanded to minimize the load on 
knee actuator $\tau_5$.
Throughout the trace, $q_4$ sweeps a range of
configurations as the coupling and applied torques evolve
(Fig.~\ref{fig:hw_coupling}, right).
The corresponding torque trace
(left) shows $\tau_4$ and $\tau_5$ consistently trading off in accordance
with the coupling relation of Eq.~\ref{eq:coupling_torque} throughout this
range. This validates the nonlinear coupling model across configurations on hardware.

\subsection{Real System PD Tracking}
To evaluate tracking performance for fast motions, a sinusoidal foot-height
trajectory, $z(t) = z_c - A\sin(2\pi f t)$ with amplitude $A=\qty{0.05}{\metre}$
about $z_c=\qty{-0.65}{\metre}$ and frequency $f=\qty{5}{\hertz}$, is commanded on
the real leg. 
\begin{figure}[!h]
    \centering

    \includegraphics[width=0.99\linewidth]{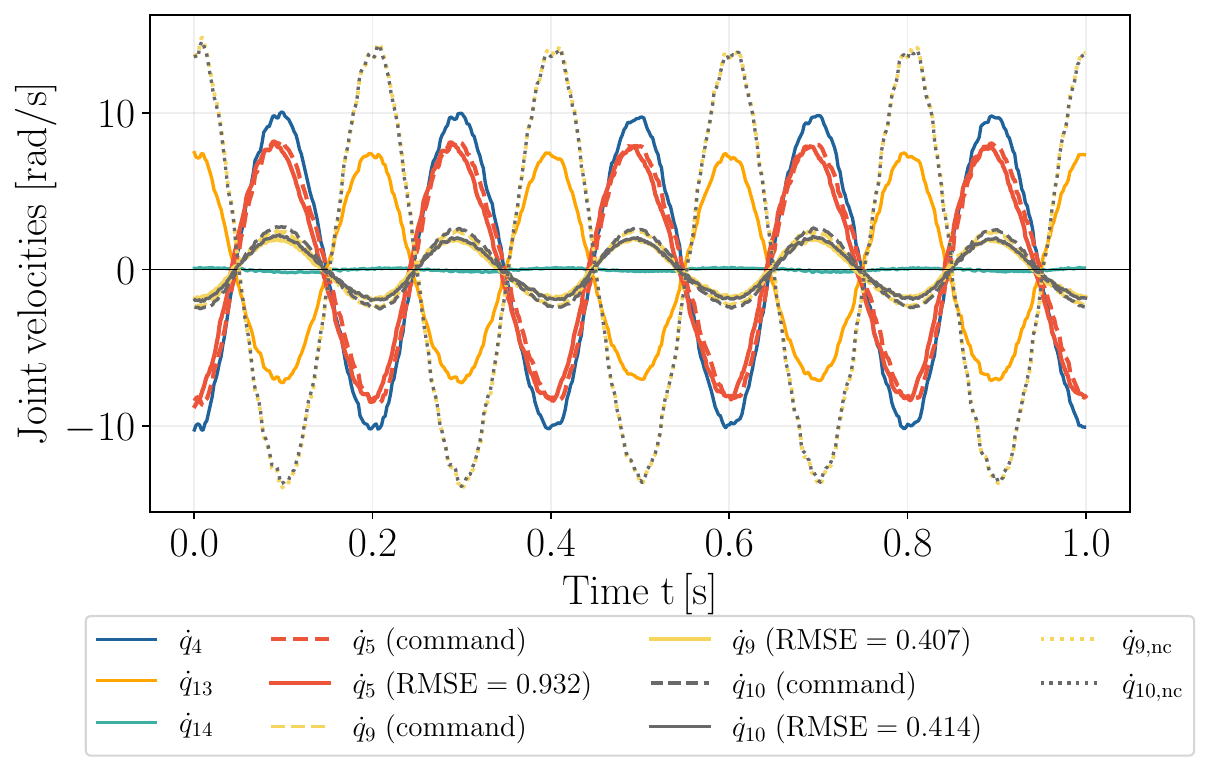}
    \caption{Velocity tracking for sinusoidal trajectory of the foot.}
    \label{fig:pd_tracking}
	\vspace{-1.0em}
\end{figure}
Fig.~\ref{fig:pd_tracking} shows the resulting velocity tracking for the knee
motor $\dot{q}_5$ and both ankle actuators $\dot{q}_9,\dot{q}_{10}$, with
per motor root mean squared error (RMSE) reported in the legend. A further observation is that the
ankle actuators show far less variation than the ankle pitch joint's own
independent velocity $\dot{q}_{13}$.  This follows directly from the coupling
Eq.~\ref{eq:coupling_velocity_ankle}, where the knee's own motion already
supplies much of the ankle actuators' required displacement. This is visible
in the sign flip where $\dot{q}_{13}$ is negative while $\dot{q}_9,\dot{q}_{10}$
are positive since it is driven by the dominant $1.6\,\dot{q}_4$ term.

To quantify this, we ask how fast the ankle actuators would need to spin
\emph{without} the coupling to produce this same ankle motion, by removing
the coupling's contribution from Eq.~\ref{eq:coupling_velocity_ankle}.
Fig.~\ref{fig:pd_tracking} overlays the result without the
coupling (dotted lines with \textit{nc} legend) where the ankle actuators would need to reach
\SI{14.82}{\radian\per\second} and \SI{14.58}{\radian\per\second}. With the
coupling, they only reach \SI{1.99}{\radian\per\second} and
\SI{2.22}{\radian\per\second} which is a \SI{86.6}{\percent} and
\SI{84.8}{\percent} reduction in peak velocity. In other words, the coupling
lets the ankle actuators do the same job at roughly a seventh of the speed an
uncoupled design would need.
The overall tracking accuracy is consistent with the leg's proximal actuator
placement. It reduces the swing phase
inertia the actuators must overcome and lets the PD controller track fast
foot motions accurately.

\subsection{Reflected Mass Model}
We compare the dynamic performance of our leg to the similar leg designs of
the MIT Humanoid~\cite{mit_humanoid} and Kangaroo~\cite{MINGOHOFFMAN2024104827}.
Two complementary quantities are evaluated over 5000 configurations sampled
uniformly within each robot's own joint limits and excluding near-singular poses. 
The swing-phase moment of inertia about the hip-pitch axis,
$I_{\text{hip}}$, sets the energetic cost of retracting and advancing the leg
each stride. The Mean Reflected Mass (MRM) $\bar m_{\hat n}$, computed from
Eq.~\ref{eq:reflected_mass} presents the average apparent inertia at the foot across all
Cartesian directions. It bounds how quickly the foot can be accelerated and
thus the leg's force-control bandwidth during stance. We focus on
$\bm{\Lambda}_v$ since
translational dynamics dominate foot-level impact and stance-phase
force control.

\begin{figure}[!h]
    \centering
    \import{figures/}{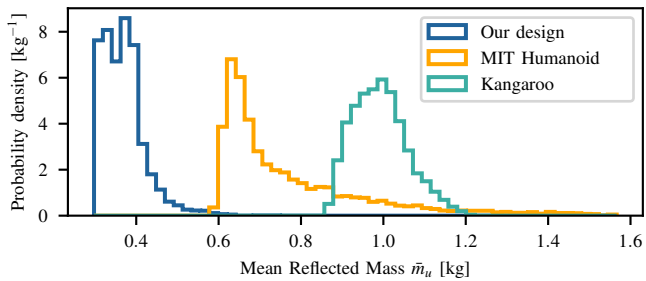}
    \caption{Mean reflected mass distribution for randomly sampled configurations in the legs' configuration spaces.}
    \label{fig:mrm_distribution}
	\vspace{-1.5em}
\end{figure}

Because the three legs differ substantially in mass $m_{\text{leg}}$ and
length $L_{\text{leg}}$ (Tab.~\ref{tab:reflected_mass_inertia}), we normalize
both quantities before comparing them. 
Without normalization, the raw MRM values (Fig.~\ref{fig:mrm_distribution}) simply
track leg mass. Our design sits around \qtyrange{0.3}{0.4}{\kg}, MIT leg around
\qtyrange{0.6}{0.8}{\kg}, and Kangaroo leg
around \qtyrange{0.9}{1.1}{\kg}. Once normalized, all three legs have nearly
the same swing inertia (0.054, 0.053, and 0.057). Their normalized MRM does
not match, with ours being the lowest at 0.061. 
Hence, our design also keeps foot-level
reflected mass low, without giving up swing-phase efficiency comparable with
current leg designs.
\begin{table}[t]
  \centering
  \caption{Leg length, mass, normalized swing inertia about the hip-pitch axis, and normalized 
  MRM at the foot}
  \label{tab:reflected_mass_inertia}
  \tabcolsep=0.12cm
  \begin{tabular}{lcccc}
    \toprule
    Robot & $L_{\mathrm{leg}}$ [m] & $m_{\mathrm{leg}}$ [kg] & $I_{\mathrm{hip}} / (m_{\mathrm{leg}} L_{\mathrm{leg}}^2)$ & $\bar{m}_{\hat{n}} / m_{\mathrm{leg}}$ \\
    \midrule
    Our design & 0.722 & 6.07 & 0.054 $\pm$ 0.017 & 0.061 $\pm$ 0.008 \\
    MIT Humanoid & 0.711 & 5.33 & 0.053 $\pm$ 0.015 & 0.145 $\pm$ 0.035 \\
    Kangaroo & 1.034 & 13.18 & 0.057 $\pm$ 0.017 & 0.075 $\pm$ 0.005 \\
    \bottomrule
  \end{tabular}
  \vspace{-1.5em}
\end{table}


\section{Conclusion and Outlook}
\label{sec:conclusion} 
This paper presented a coupled series-parallel transmission leg, combining
a knee parallel mechanism, an ankle parallel mechanism, and a nonlinear
knee-ankle coupling where all actuators are placed closer to the hip. We derived
the coupling's nonlinear transmission models and quantified its
mechanical advantage at both the torque and velocity level.
The coupling eases the knee's peak torque demand under load and the
ankle's peak speed demand during fast motion. This lets both joints draw on
each other's actuator capacity rather than acting independently.
We verify these relations directly on hardware. With the leg's proximal 
actuator placement, the resulting leg achieves an effective reflected 
mass of \qty{0.37}{\kg} at the foot for a \qty{6}{\kg} leg, a very low distal leg inertia 
for dynamic humanoid robots.

Future work includes integrating this leg into a full humanoid and
exploiting the dynamic capabilities enabled by the proposed coupled
transmissions. Further experiments will be performed to identify
belt backlash and friction in the nonlinear coupling.

%

\bibliographystyle{IEEEtran}
\bibliography{references}

\end{document}